# Optimal Technical Indicator-based Trading Strategies Using NSGA-II

*P. Shanmukh Kali Prasad[1], Vadlamani Madhav[2], Ramanuj Lal[3] and Vadlamani Ravi[4*]*

[1]*Birla Institute of Technology and Science, Pilani, Hyderabad Campus- 500078*

[2]*Department of Electrical Engineering, IIT Bombay, Powai, Mumbai-400076*

[3]*QR Systems, Mumbai, India*

[4]*Centre of Excellence in Analytics,
Institute for Development and Research in Banking Technology,
Castle Hillas Road #1, Masab Tank, Hyderabad-500057*

shanmukh.kp@gmail.com, vadpadam@gmail.com, rav_padma@yahoo.com, ramanujlal@gmail.com

**Abstract**

This paper proposes non-dominated sorting genetic algorithm-II (NSGA-II ) in the context of technical indicator-based stock trading, by finding optimal combinations of technical indicators to generate buy and sell strategies such that the objectives, namely, *Sharpe ratio and Maximum Drawdown* are maximized and minimized respectively. NSGA-II is chosen because it is a very popular and powerful bi-objective evolutionary algorithm. The training and testing used a rolling-based approach (two years training and a year for testing) and thus the results of the approach seem to be considerably better in stable periods without major economic fluctuations. Further, another important contribution of this study is to incorporate the transaction cost and domain expertise in the whole modeling approach.

1. Introduction

The NSGA-II algorithm was proposed in 2002 [1] and was praised for its lower computational complexity and elitist approach. In this paper, we incorporate this algorithm into the world of technical analysis in finance. Technical analysis refers to the study of analysing patterns through the usage of technical indicators. Traders who favour technical analysis are known to prefer combinations of technical indicators over individual indicators and that is the realm this paper seeks to venture into.

---

[*] Corresponding Author; Phone: +914023294310; FAX: +914023535157

The usage of genetic algorithms in trading using technical indicators has been recorded earlier in works such as references [2] [3]. These works formulated an optimization problem where a single objective was considered to maximize. This maximization was done by evaluating randomly generated strategies (viz., combinations of technical indicators) until the best ones were found. Literature in the usage of genetic algorithms in this domain considered the Sterling ratio to be a popular objective. The Sterling ratio is a ratio between annualized returns and the Maximum Drawdown and is well known in financial literature.

Our work formulates a bi-objective optimization problem and seeks to generate strategies that use combinations of technical indicators to maximise these objectives. While there are recorded works of evolutionary computational techniques in portfolio optimization problems such as [4] and [5], the problems they formulated were the standard risk-minimizing, return-maximizing approach. Instead of following this standard approach, we construct a different optimization problem where we seek to maximise Sharpe Ratio and minimise Maximum Drawdown using the NSGA-II algorithm. The methodology involves a rolling training and testing approach (two years training, followed by testing on the third year) to ensure that the contribution of irrelevant market conditions to strategy generation is minimal. This approach allows the generation of unique strategies where investors can have the best returns over periods of time while also minimizing the maximum losses that they face in the investment horizon. Apart from focussing on a bi-objective problem without much precedence, this paper also took the help of a domain expert and considered two different avenues, namely- classifying indicators into groups while generating strategies and incorporating transaction costs. Grouping indicators was included to enforce certain constraints upon the strategies being generated to enhance their digestibility in practical applications by the industry. Transaction costs were incorporated by assuming a percentage cost for every position taken in the simulated market during back testing.

This paper is organized as follows; Section 2 presents the literature review of the evolutionary computational algorithms applied to in various problems of the finance domain. This is followed by a section 3 that explains the proposed methodology that includes the solution encoding scheme adopted here as well as the details of the fitness function evaluation.

Then section 4 discusses the empirical results of the strategies obtained over the investment horizon are described and finally section 5 concludes the paper.

## 2. Literature Review

Of late, the usage of evolutionary computational algorithms in the finance sector has been a competitive sector of research. Literature surveys such as [6], [7] indicate a rise in popularity of evolutionary computation techniques to enhance both fundamental and technical analysis in the financial trading sector. Among the various techniques found in the field of evolutionary computing today, a particular preference to the usage of genetic algorithms and genetic programming in the financial domain was observed by these surveys. Another literature survey [8] demonstrates a growing interest in incorporating evolutionary techniques in the reinforcement learning domain as well.

Stock analysis and investment decisions are made through either of the two approaches- fundamental or technical. Fundamental analysis comprises of rigorous observations made using book values, financial ratios of the company and macroeconomic analyses. Technical analysis involves quick decision-making by observing market opinions towards a financial instrument using technical indicators. A literature survey [6] demonstrates a strong preference towards the usage of evolutionary computational methods in technical analysis rather than fundamental one. However, works such as [9] and [10] have shown impressive developments in the fundamental analysis field through the usage of fuzzy genetic programming and genetic algorithms respectively. Ref. [9] involved fuzzy genetic programming to build multi-valued stock valuation models and demonstrated a trading strategy that worked significantly better than a simple buy-and-hold strategy, Ref. [10] described an approach to parallelize genetic algorithms using intra-day data and indicators used in fundamental analysis (macroeconomic variables, industry indicators etc.) to come up with strategies that considerably outperformed the S&P 500.

Research in the usage of evolutionary computation in the field of blending analysis (analysis using both technical and fundamental methods) is a growing area. Works like Ref. [11] involved using a type-2 fuzzy system on fundamental and technical indicators to make robust stock predictions.

Recent work by Huang et.al. [12] proposed a GA-based system that used stock price data at microscopic levels collected at high frequencies to significantly improve price movement prediction models. Evolutionary computational algorithms are being increasingly preferred to complex deep-learning models in financial forecasting or prediction problems because of the interpretability they can offer. [13] This interpretability has allowed prediction problems to develop into unique approaches where generalized rules could be identified using results from these algorithms. For instance, work by Tsang et.al. [14] -which involved training a GP using past data from financial stock markets to predict price movements- allowed the authors to identify the most successful indicators. Other work by Allen et.al. [15] allowed the authors to discover new financial indicators through the usage of genetic algorithms. Thus, research in this area has led to the creation a vast number of approaches to generate multiple trading rules that can be chosen by investors based on their risk preferences. An example of research such has this was done by -Almanza et.al. [16] where "Evolving Decision Rules" were introduced using an evolutionary algorithm and a fuzzy logic rule-based representation. These results included examples of experimental sets of rules that could fit the risk guidelines of different types of investors.

Another area of evolutionary computation research in the finance domain besides forecasting and prediction is in portfolio optimization problems. Portfolio optimization opens doors to multi-objective optimization problem formulation and thus, a multitude of approaches can be used to tackle this. Two of the objectives very often considered in a portfolio optimization are return and risk, where the former is maximized, and the latter is minimized. Work by Hassan et.al. [5] accomplishes precisely this and incorporates a robustness measure into a multi-objective genetic programming fitness function. Diosan [4] also considered risk and return (profit) as objectives and compared the results of three evolutionary algorithms (NSGA-II, PESA and SPEA2) in the work..

Another area of research in the finance domain where evolutionary computation makes an appearance is in option pricing. While the Black Scholes model prices are used as references even today, certain assumptions that aren't always representative of real-life conditions cause departures from the theoretical values. Evolutionary algorithms have been used to perform pricing in several works such as [17] and [18]. Ref. [17] showed promising results even though the Black Scholes model performed better. Ref. [18] was one of the first applications of quantum inspired evolutionary algorithms in the financial domain. While its results weren't remarkable, it shows scope for several methodologies of research.

As mentioned earlier in this section, evolutionary algorithms are being increasingly used to generate trading strategies because of the interpretability of their results. Dempster and Jones [2] developed a trading system for forex markets using high frequency foreign exchange tick data from 1994 to 1997. They adopted genetic programming procedures and used technical indicators to generate rules that best optimised their single objective- a modified version of the Sortino ratio. They concluded that their best results yielded around 7% annualised returns but were, however, mostly loss-making strategies. This paper however was impactful in illustrating the usage of genetic algorithms in technical trading.

Hryshko and Downs [3] later demonstrated the use of a hybridized GA-RL model in foreign exchange markets and high frequency forex data by using a genetic algorithm model like the one mentioned above and then using a Q-Learning based RL engine They achieved moderate gains in their out-sample testing periods. Another usage of hybridised GA-RL models was reported by Zhang et al., [19] where recurrent reinforcement learning was used in an equity training system with the dataset being 180 S&P stocks' daily data. The trading system again employed genetic algorithms to achieve optimal technical indicator rules that enabled maximising of the Sharpe Ratio. They succeeded in demonstrating results where the Sharpe ratios were significantly greater than zero. Dempster and Romahi [20] worked on hybridized GA-RL models too and demonstrated better results than those of a standard RL model as well..

Briza et al., [21] used another evolutionary computation algorithm, that supported multiple objectives called the multiple objective particle swarm optimization (MOPSO) on end-of-day market data. The objectives they sought to maximize were percentage profit and Sharpe Ratio. Instead of using Boolean technical indicator-based chromosomes like most papers mentioned above, he adopted a weightage-based approach to assign technical indicators importance. This method helped generate Pareto fronts and was said to outperform NSGA II significantly. Maximum Drawdown was suggested as a potential objective in the future.

3. **Proposed Methodology**

**3.1 Brief overview of Non-dominated Sorting Genetic Algorithm-II (NSGA-II)**

NSGA-II [1] is an advanced multi-objective optimization algorithm, which uses a sorting algorithm, incorporates elitism, and does not require niching or any other parameter. In NSGA-II, a population is randomly initialized. Then, the population is sorted based on non-domination of solutions resulting in various hierarchical fronts. Once the non-dominated sorting is completed, a crowding distance value is assigned to each solution depending on its closeness to other population members in the objective space. The individuals are selected based on their non-dominated rank and the crowding distance, in that order. Thereafter, crossover and mutation operators are applied to the selected parent solutions to create the new offspring population. This process continues until a termination criterion is reached. At every subsequent generation t, the offspring population $Q_t$ and the current generation population $P_t$ are combined and selection is invoked to select the solutions for the next generation. Elitism is taken care of because all the previous and current best solutions are passed on to the next-generation. Then, the population is sorted based on non-domination criterion. Thereafter, new generation is formed by each front until the population size exceeds the current population size. In case the addition of all the solutions in the front $F_i$ results in the population $P_{t+1}$ exceeding N, then solutions in front $F_i$ are selected based on their crowding distance in the descending order until the population size is N. The process is repeated in the subsequent generations. This schematic is depicted in Fig. 1.

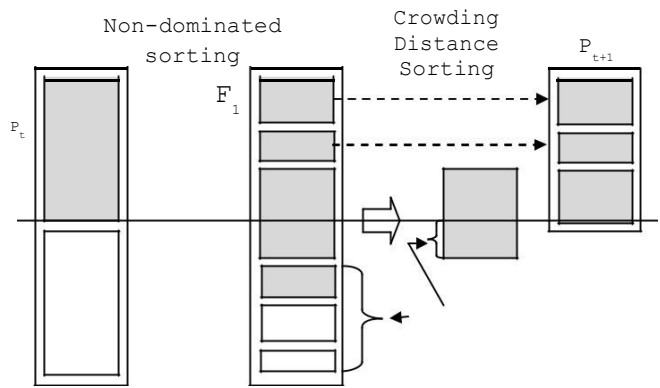

Figure 1: Non-dominated Sorted Genetic Algorithm-II (NSGA-II)

## 3.2 Proposed Approach

First, popular technical indicators known in the technical analysis domain are calculated with the OHLC data. Following this, signals (Buy, Sell and Hold) are generated for each of these indicators using rules (refer- Appendix) recommended by domain experts in the trading sector. For example, consider the signals generated by using the rules from the Price Oscillator indicator-

$$\text{Price Oscillator} = \frac{(10\text{Day EMA of Closing Prices} - 20\text{Day EMA of Closing Prices})}{20\text{Day EMA of Closing Prices}}$$

-where EMA is exponential moving average.

Rules:

PO_BUY=1, if PO(n-1) <0 and PO(n)>0

PO_SELL=1, if PO(n-1) >0 and PO(n)<0 where n refers to the nth day.

After all the signals are generated, binary encoded strings called chromosomes are constructed where each bit is referred to as a gene. These chromosomes function as a Boolean representation of the final strategies. The structure of the chromosome will be explained in the next section.

Following the generation of the signals, chromosomes are generated randomly (with a few constraints) to create an initial population. This initial population goes through the NSGA-II algorithm and undergoes operations like crossover, mutation and selection before a final pareto front of strategies is generated. NSGA-II incorporates a fast non-dominated sorting and elitist selection as a part of its optimization exercise. In a bi-objective problem like the current one, a solution $x_1$ is said to pareto dominate another point $x_2$ if the below conditions hold [22]-

$$f_i(x_1) \geq f_i(x_2) \ \forall \ i \ \in (1,2)$$

$$f_i(x_1) > f_i(x_2) \ for \ at \ least \ i \ such \ that \ i \ \in (1,2)$$

where $f_i$ represents the $i^{th}$ objective function. (Sharpe ratio or Maximum Drawdown)

## 3.3 Encoding of the solution

Our study considers 9 technical indicators like those used by [3] while trying to figure out which among these form the most optimal combinations. With the help of a domain expert, we have classified these 9 indicators into two groups- namely, momentum and reversal. Momentum indicators try to check whether the current trend in the market continues, and reversal indicators check to see whether these trends have reached their limit and are about to reverse their direction. In the order mentioned in the appendix, we classified the first four indicators as momentum and the rest as reversal. As mentioned earlier in the previous section, all these indicators have their own rules to determine when a buy/sell signal is generated.

Now, we define the chromosome's structure. All chromosomes are binary strings of a fixed length of 52. The first 26 genes represent the BUY rule, and the next represent the SELL rule. Each of these halves can again be divided as a string of 17 and 9. The latter 9 simply represents which indicators are active in our rule i.e., which indicators are being considered. It has no say on whether the signal given by the indicator necessarily must be TRUE. Now, the 17-bit string can be divided into a 9-bit string and 8 connectors (AND or OR). This 9-bit string represents whether the indicators which are active have a necessary TRUE/FALSE value. It may be easier to understand this structure with the following example.

[0, 0, 0, 0, 1, 0, 1, 1, 0, 0, 0, 0, 0, 0, 1, 0, 0,0, 1, 1, 1, 0, 1, 1, 1, 1, 0, 0, 0, 0, 0, 0, 1, 1,0, 0, 0, 0, 1, 0, 0, 0, 0, 1, 1, 1, 1, 1, 0, 1, 0, 0]

Given above is an example of a chromosome. We'll try breaking this chromosome into a strategy of IF ELSE rules.

Let's divide the chromosome first into four strings: Buy Structure, Buy Indicators, Sell Structure, Sell Indicators.

- Buy Structure: First 17 genes.
  [0, 0, 0, 0, 1, 0, 1, 1, 0, 0, 0, 0, 0, 0, 1, 0, 0]
- Buy Indicators: Next 9 genes.
  [0, 1, 1, 1, 0, 1, 1, 1, 1]
- Sell Structure: Next 17 genes.
  [0, 0, 0, 0, 0, 0, 1, 1,0, 0, 0, 0, 1, 0, 0, 0, 0]
- Sell Indicators: Last 9 genes.
  [1, 1, 1, 1, 1, 0, 1, 0, 0]

From the Buy Indicators string we can say that MACD, MO, PO, RSI, CCI, LW, BB are active indicators in the buy structure.[1] Now looking at the buy structure, we can formulate a rule.

*IF ((MACD_BUY=FALSE & MO_BUY=TRUE & PO_BUY=TRUE) OR (RSI_BUY=FALSE & CCI_BUY=FALSE & LW=1 & BB_BUY=FALSE)):*

    *BUY_SIGNAL=TRUE*

*IF ((SMA_SELL=FALSE & MACD_SELL=FALSE & MO_SELL=FALSE & PO_SELL=TRUE) OR (SO_SELL=FALSE AND CCI_SELL=TRUE))*

    *SELL_SIGNAL=TRUE*

*IF BUY_SIGNAL=TRUE & SELL_SIGNAL=FALSE:*

    *FINAL_SIGNAL= 1*

---

[1] The order of the indicators in the string is same as the order of the indicators in the Appendix section.

*IF SELL_SIGNAL=TRUE & BUY_SIGNAL=FALSE:*

    *FINAL_SIGNAL= -1*

*IF BUY_SIGNAL=SELL_SIGNAL:*

    *FINAL_SIGNAL= 0*

After calculating technical indicators for each day in the dataset, and generating the signals, check to see if the signals match the conditions stated by the chromosome and generate BUY or SELL signals accordingly.

### 3.4 Fitness Function Evaluation

It is important to evaluate how effective a strategy obtained from a chromosome is. As mentioned earlier, our goal is to maximise Sharpe Ratio and minimize Maximum Drawdown. Sharpe Ratio is the ratio between annualized portfolio return and annualized volatility (standard deviation of portfolio returns). Maximum Drawdown is the maximum loss endured during the trading time. The Sharpe ratio is an important tool in evaluating the risk-return trade-off while the Maximum Drawdown focusses on preserving capital. It is important to note however that the latter objective gives no idea about the frequency of such losses and therefore must be treated with caution. The procedure for compting the Sharpe Ratio and Maximum Drawdown is as follows.

### 3.4.1 Sharpe Ratio Calculation

The Sharpe Ratio is defined as the ratio between the annualized return and annualized volatility. It helps create an effective measure of the risk-return tradeoff investors face. Here the Sharpe Ratio is calculated after adding in a percentage for transaction costs as well.

Calculation:

Let the open, high, low close price for an asset $i$ on day $t$ be $o_{i,t}, h_{i,t}, l_{i,t}, c_{i,t}$. The long-only total return on the asset on day t can be calculated as:

$$r_{i,t} = \frac{c_{i,t} - c_{i,t-1}}{c_{i,t-1}}$$

The long-only return indicates the daily return gained by an investor if a long position was held that day.

Based on the signal we may take the following three positions in the asset in our portfolio, corresponding to "Long", "Short" or "Neutral/No position"

$$w_{i,t} = +1 \text{ or } -1 \text{ or } 0$$

It is important to note that the position we take depends on the signal we get using the closing price data at the end of the previous day. This prevents look-ahead bias. Look-ahead bias refers to the bias caused in back-testing processes when we run trades with data we should not have at that instant of time.

Based on the position we take; we earn the portfolio excess return-

$$r_{p,t}^{gross} = w_{i,t} * r_{i,t}$$

The portfolio turnover on any given day can be calculated as

$$\tau_{i,t} = w_{i,t} - w_{i,t-1} * \frac{1 + r_{i,t-1}}{1 + r_{p,t-1}^{gross}}$$

The portfolio excess return net i.e., the return obtained after deducting transaction costs-

$$r_{p,t}^{net} = r_{p,t}^{gross} - \tau_{p,t} * k$$

Here, k is assumed to be a constant transaction cost. The second term above is the total transaction cost that we are paying. The turnover represents a weighted change in our position (long/short/neutral).

The annualized total return of the portfolio is given by taking the geometric mean of the returns of all our trades made in the stipulated time.

$$R_p^{net} = \left( \prod_t (1 + r_{p,t}^{net}) \right)^{252/N} - 1$$

Where N is the total number of days. 252 represents the number of functional days of a stock market per annum.

The annualized volatility of the portfolio return is the standard deviation of the returns of our trades multiplied with a factor of 16. The factor 16 represents the equivalent of the root of 252 and is used for annualizing our daily return volatility. This volatility acts as variable denoting risk.

$$\Sigma_p = 16 * \sqrt{(\sum \frac{(r_{i,t} - \overline{r_{i,t}})^2}{N})}$$

The Annualized Sharpe Ratio is defined as

$$S = \frac{R_p^{net}}{\Sigma_p}$$

**3.4.2 Maximum Drawdown Calculation**

To calculate Maximum Drawdown, we first calculate a wealth index. The wealth index acts as a representation of our portfolio value regardless of the capital initially invested.

The wealth index on a Day-X is obtained by taking the cumulative product of returns of all days from Day-0 to Day-X.

$$WI_x = \prod_{i=0}^{x}(1 + r_i)$$

The next step is to calculate the highest peaks of the wealth indices that occurred from Day-0 to Day X for each Day X.

$$Peak_x = \max(WI_0, WI_1 \ldots WI_x)$$

Once the peaks are calculated, the drawdown of each day is the percentage change between the wealth index of each day and the highest peak till that day.

$$Drawdown_x = 100 * \frac{WI_x - Peak_x}{Peak_x}$$

The lowest of these differences is the Maximum Drawdown.

## 4. Data description

The datasets used were daily OHLC (Open, High, Low, Close) prices of renowned stock indices like BSE Sensex and NIFTY 50 obtained from Yahoo Finance[2] for 20-year time periods. The idea was to consider indices over individual stocks to account for diversification of portfolios and focussing on systematic risk. The data collected however was not divided into lumps of training and testing periods. A rolling window-based approach was implemented instead to train and test in 2-year and 1-year intervals respectively.

## 5. Results and discussion

The following results were obtained using a population size of 30, a crossover rate of 0.9, a mutation rate of 0.1 and a maximum of five generations. The entire simulation was written using Python on Jupyter IDE and was run on a machine with a quad-core Intel i5-11th generation 2.40 GHz processor.

The results obtained from training and testing using a rolling-based mechanism are presented below. The In-Sample and Out-Sample results refer to the training and testing periods respectively. It is important to note that the results shared below in each table form a Pareto front i.e., none of the points can be said to dominate any of its peers. For each of the results, the buy and sell strategy columns represent our conditions to take up long and short positions in the market respectively. As mentioned before in the Encoding of the Solution section, one important point to notice among the strategies is the set of constraints enforced. The AND operator was fixed between all the technical indicators of each category. The two categories of indicators then form one strategy using a fixed OR operator.

Based on OHLC SENSEX data from 2005-15, at a transaction cost of 2%, we get the following results.

---

[2] https://finance.yahoo.com/

**Table 1. Strategies in In-Sample Period: 2003-2004 Out-Sample Period: 2005***

| S.No | Sharpe Ratio & Drawdown (In-Sample) | Sharpe Ratio & Drawdown (Out-Sample) | Buy Strategy | Sell Strategy |
|---|---|---|---|---|
| 1 | [4.879,-0.042] | [2.275,-0.114] | IF sto_buy = 0.0 AND RSI_buy = 0.0 AND CCI_buy = 1.0 | IF SMA_sell = 0.0 AND MO_sell = 1.0 OR sto_sell = 1.0 AND CCI_sell = 1.0 AND LW_sell = 0.0 |
| 2 | [5.323,-0.046] | [1.645,-0.138] | IF SMA_buy = 0.0 AND MO_buy = 1.0 OR sto_buy = 0.0 AND CCI_buy = 0.0 | IF SMA_sell = 0.0 AND MO_sell = 1.0 AND PO_sell = 1.0 OR RSI_sell = 1.0 AND BB_sell = 0.0 |
| 3 | [5.595,-0.0469] | [1.566.-0.138] | IF SMA_buy = 1.0 AND MO_buy = 0.0 OR sto_buy = 0.0 AND RSI_buy = 0.0 AND CCI_buy = 0.0 | IF SMA_sell = 0.0 AND MO_sell = 1.0 AND PO_sell = 1.0 OR RSI_sell = 1.0 AND BB_sell = 0.0 |

* Periods of economic stability made strategies perform better in the testing period.

In Table 1, it can be determined from the in-sample column that a pareto front of three strategies was generated. However, upon testing on the out-sample data, the first strategy performed better than the rest with a higher Sharpe ratio and lower maximum drawdown (11.4%). In the first strategy, it can be observed that for the buy strategy none of the indicators from the first category (Momentum) were deemed important.

**Table 2. Strategies in In-Sample Period: 2004-2005 Out-Sample Period: 2006**

| S.No | Sharpe Ratio & Drawdown (In-Sample) | Sharpe Ratio & Drawdown (Out-Sample) | Buy Strategy | Sell Strategy |
|---|---|---|---|---|
| 1 | [3.464, -0.099] | [ 2.340, -0.190] | 'IF PO_buy = 0.0 OR sto_buy = 0.0 AND LW_buy = 1.0 AND BB_buy = 1.0' | 'IF SMA_sell = 0.0 AND MACD_sell = 0.0 AND PO_sell = 1.0 OR sto_sell = 0.0 AND RSI_sell = 0.0 AND CCI_sell = 1.0 AND BB_sell = 0.0' |
| 2 | [ 3.497, -0.111] | [ 1.812, -0.261] | IF MO_buy = 0.0 OR CCI_buy = 0.0 | IF MO_sell = 0.0 AND PO_sell = 1.0 OR sto_sell = 1.0 AND RSI_sell = 1.0 AND CCI_sell = 0.0 AND LW_sell = 0.0 AND BB_sell = 1.0 |
| 3 | [3.508,-0.1192] | [ 2.162, -0.257] | IF SMA_buy = 0.0 AND PO_buy = 0.0 OR sto_buy = 1.0 AND RSI_buy = 0.0 AND CCI_buy = 1.0 AND LW_buy = 0.0 | IF SMA_sell = 0.0 AND MACD_sell = 0.0 AND MO_sell = 0.0 AND PO_sell = 1.0 OR sto_sell = 1.0 AND RSI_sell = 1.0 AND BB_sell = 0.0 |

In Table 2, a pareto front of three strategies was generated. In contrast to the previous investment horizon, this time frame shows a good distribution of technical indicators of both categories in the buy and sell strategies.

**Table 3. Strategies in In-Sample Period: 2005-2006 Out-Sample Period: 2007**

| S.No | Sharpe Ratio & Drawdown (In-Sample) | Sharpe Ratio & Drawdown (Out-Sample) | Buy Strategy | Sell Strategy |
|---|---|---|---|---|
| 1 | [0.528, -0.046] | [-1.002, -0.081] | IF MACD_buy = 1.0 AND MO_buy = 1.0 AND PO_buy = 1.0 OR sto_buy = 1.0 AND RSI_buy = 1.0 AND BB_buy = 1.0 | IF SMA_sell = 0.0 AND PO_sell = 1.0 OR sto_sell = 1.0 AND RSI_sell = 0.0 AND CCI_sell = 1.0 AND LW_sell = 1.0 AND BB_sell = 1.0 |
| 2 | [0.816, -0.054] | [1.283, -0.059] | IF MO_buy = 1.0 OR sto_buy = 1.0 | IF SMA_sell = 0.0 AND MO_sell = 1.0 AND PO_sell = 1.0 OR BB_sell = 1.0 |
| 3 | [2.145, -0.179] | [1.157, -0.168] | IF SMA_buy = 1.0 AND MACD_buy = 0.0 AND MO_buy = 0.0 OR RSI_buy = 0.0 AND CCI_buy = 0.0 | IF SMA_sell = 1.0 AND MACD_sell = 1.0 AND MO_sell = 0.0 AND PO_sell = 0.0 OR sto_sell = 0.0 AND RSI_sell = 0.0 AND CCI_sell = 1.0 |
| 4 | [2.803, -0.225] | [1.626, -0.175] | 'IF SMA_buy = 0.0 OR RSI_buy = 1.0 | IF SMA_sell = 1.0 AND MACD_sell = 1.0 AND MO_sell = 0.0 AND PO_sell = 0.0 OR sto_sell = 0.0 AND RSI_sell = 0.0 AND CCI_sell = 1.0 |

In Table 3, a pareto front of four strategies was generated. The first strategy generated here has the lowest maximum drawdown but also has an unsatisfactory Sharpe Ratio. This strategy fails to perform well in the out-sample data as well. It is a prime example of making necessary judgement calls and evaluating the strategies effectively before their usage.

**Table 4. Strategies in In-Sample Period: 2006-2007 Out-Sample Period: 2008**

| S.No | Sharpe Ratio & Drawdown (In-Sample) | Sharpe Ratio & Drawdown (Out-Sample) | Buy Strategy | Sell Strategy |
|---|---|---|---|---|
| 1 | [0.955, -0.059] | [-0.885, -0.147] | IF MACD_buy = 0.0 AND MO_buy = 1.0 AND PO_buy = 0.0 OR sto_buy = 1.0 AND RSI_buy = 1.0 AND LW_buy = 0.0 AND BB_buy = 0.0 | IF SMA_sell = 0.0 AND MACD_sell = 0.0 AND MO_sell = 1.0 AND PO_sell = 1.0 OR RSI_sell = 1.0 AND CCI_sell = 0.0 AND BB_sell = 1.0 |
| 2 | [0.993, -0.088] | [-0.379, -0.132] | IF MO_buy = 1.0 OR CCI_buy = 0.0 AND LW_buy = 1.0 | IF SMA_sell = 1.0 AND MACD_sell = 1.0 AND PO_sell = 0.0 OR BB_sell = 1.0 |
| 3 | [2.100, -0.226] | [-1.660, -0.624] | 'IF SMA_buy = 0.0 AND MO_buy = 1.0 OR sto_buy = 0.0 AND RSI_buy = 0.0' | IF MACD_sell = 1.0 AND PO_sell = 0.0 OR sto_sell = 1.0 AND CCI_sell = 1.0' |
| 4 | [2.112, -0.269] | [-1.423, -0.568] | 'IF SMA_buy = 0.0 AND MO_buy = 0.0 OR sto_buy = 0.0 AND RSI_buy = 0.0' | 'IF MACD_sell = 1.0 AND PO_sell = 0.0 OR sto_sell = 1.0 AND CCI_sell = 1.0' |

In Table 4, the investment horizon is a prime example of severe economic fluctuations and thus, the strategies do not give results as desired.

**Table 5. Strategies in In-Sample Period: 2007-2008 Out-Sample Period: 2009**

| S.No | Sharpe Ratio & Drawdown (In-Sample) | Sharpe Ratio & Drawdown (Out-Sample) | Buy Strategy | Sell Strategy |
| --- | --- | --- | --- | --- |
| 1 | [0.336, -0.042] | [0.497, -0.021] | IF sto_buy = 1.0 AND RSI_buy = 1.0 AND CCI_buy = 0.0 | IF PO_sell = 0.0 OR CCI_sell = 1.0 AND LW_sell = 0.0 |
| 2 | [0.731, -0.108] | [0.558, -0.102] | IF SMA_buy = 0.0 AND PO_buy = 1.0 OR sto_buy = 0.0 AND RSI_buy = 0.0 AND CCI_buy = 0.0 AND LW_buy = 1.0 | IF MO_sell = 1.0 OR CCI_sell = 0.0 AND LW_sell = 1.0 |

In Table 5, a pareto front of two strategies was generated. The results in the in-sample and out-sample periods are exceedingly similar and show the effectiveness of using the generated strategies in stable or consistent economic conditions.

**Table 6. Strategies in In-Sample Period: 2008-2009 Out-Sample Period: 2010**

| S.No | Sharpe Ratio & Drawdown (In-Sample) | Sharpe Ratio & Drawdown (Out-Sample) | Buy Strategy | Sell Strategy |
| --- | --- | --- | --- | --- |
| 1 | [1.356, -0.044] | [-0.52, -0.064] | IF MACD_buy = 0.0 OR sto_buy = 0.0 AND BB_buy = 0.0 | IF MACD_sell = 1.0 AND PO_sell = 1.0 OR sto_sell = 0.0 AND RSI_sell = 0.0 |

In Table 6, only one strategy was generated as a part of the Pareto front i.e., a single strategy proved to be better than all the other strategies in the in-sample data. Indicators from both classes such as MACD and Stochastic Oscillator were deemed important. However, this strategy doesn't perform well in the out-sample data and is an indicator of changing economic situations.

**Table 7. Strategies in In-Sample Period: 2009-2010 Out-Sample Period: 2011**

| S.No | Sharpe Ratio & Drawdown (In-Sample) | Sharpe Ratio & Drawdown (Out-Sample) | Buy Strategy | Sell Strategy |
|---|---|---|---|---|
| 1 | [ 0.413, -0.037] | [-0.172, -0.065] | 'IF MACD_buy = 0.0 OR RSI_buy = 1.0 AND LW_buy = 1.0 | IF OR sto_sell = 0.0 AND CCI_sell = 1.0 AND BB_sell = 0.0 |
| 2 | [ 0.69, -0.039] | [ 0.033, -0.069] | IF SMA_buy = 1.0 OR sto_buy = 0.0 AND RSI_buy = 0.0 AND CCI_buy = 1.0 AND LW_buy = 0.0 AND BB_buy = 1.0 | 'IF SMA_sell = 0.0 AND MACD_sell = 0.0 AND MO_sell = 0.0 AND PO_sell = 1.0 OR sto_sell = 1.0 AND LW_sell = 1.0 AND BB_sell = 1.0' |
| 3 | [ 0.757, -0.053] | [-0.327, -0.095] | 'IF MACD_buy = 0.0 AND PO_buy = 1.0 OR RSI_buy = 0.0 AND LW_buy = 1.0 | IF SMA_sell = 1.0 AND MACD_sell = 0.0 AND MO_sell = 1.0 OR sto_sell = 0.0 AND LW_sell = 0.0 AND BB_sell = 1.0' |
| 4 | [ 2.597, -0.125] | [-0.672, -0.194 ] | 'IF MACD_buy = 0.0 OR RSI_buy = 1.0 AND LW_buy = 1. | IF SMA_sell = 0.0 AND MO_sell = 1.0 AND PO_sell = 1.0 OR RSI_sell = 0.0 AND CCI_sell = 1.0 AND BB_sell = 0.0 |

Despite a larger Pareto front, the results in Table 7 are quite similar to Table 6 in terms of indicator distribution and performance.

**Table 8. Strategies in In-Sample Period: 2010-11 Out-Sample Period: 2012**

| S.No | Sharpe Ratio & Drawdown (In-Sample) | Sharpe Ratio & Drawdown (Out-Sample) | Buy Strategy | Sell Strategy |
|---|---|---|---|---|
| 1 | [0.489, -0.029] | [-1.410, -0.021] | IF SMA_buy = 1.0 AND MACD_buy = 1.0 AND MO_buy = 0.0 AND PO_buy = 1.0 OR RSI_buy = 1.0 AND LW_buy = 1.0 AND BB_buy = 1.0' | IF SMA_sell = 1.0 OR sto_sell = 1.0 AND CCI_sell = 1.0 AND LW_sell = 0.0 |
| 2 | [0.544, -0.036] | [-0.495, -0.034] | F SMA_buy = 1.0 AND PO_buy = 0.0 OR RSI_buy = 1.0 AND LW_buy = 1.0 AND BB_buy = 1. | IF SMA_sell = 1.0 AND MACD_sell = 0.0 AND MO_sell = 1.0 AND PO_sell = 0.0 OR RSI_sell = 1.0 AND CCI_sell = 0.0 AND LW_sell = 1.0 |

In Table 8, a pareto front of two strategies was generated. This time frame shows a good distribution of technical indicators of both categories in the buy and sell strategies.

**Table 9. Strategies in In-Sample Period: 2011-2012 Out-Sample Period: 2013**

| S.No | Sharpe Ratio & Drawdown (In-Sample) | Sharpe Ratio & Drawdown (Out-Sample) | Buy Strategy | Sell Strategy |
|---|---|---|---|---|
| 1 | [0.954, -0.049] | [0.260, -0.078] | IF MACD_buy = 0.0 AND MO_buy = 1.0 AND PO_buy = 0.0 OR RSI_buy = 0.0 AND CCI_buy = 0.0 AND LW_buy = 1.0 AND BB_buy = 1.0 | IF MO_sell = 1.0 AND PO_sell = 0.0 OR sto_sell = 0.0 AND RSI_sell = 1.0 AND CCI_sell = 0.0 AND LW_sell = 0.0 AND BB_sell = 0.0 |

In Table 9, only one strategy was generated as a part of the Pareto front i.e., a single strategy proved to be better than all the other strategies in the in-sample data with respect to both the objectives considered (Sharpe Ratio and Maximum Drawdown).

**Table 10. Strategies in In-Sample Period: 2012-2013 Out-Sample Period: 2014**

| S.No | Sharpe Ratio & Drawdown (In-Sample) | Sharpe Ratio & Drawdown (Out-Sample) | Buy Strategy | Sell Strategy |
|---|---|---|---|---|
| 1 | [1.190, -0.106] | [2.595, -0.057] | IF RSI_buy = 1.0 AND LW_buy = 1.0 AND BB_buy = 1.0 | IF MACD_sell = 1.0 OR sto_sell = 1.0 AND LW_sell = 1.0 |
| 2 | [1.284, -0.107] | [2.556, -0.055] | IF OR RSI_buy = 0.0 AND LW_buy = 1.0 AND BB_buy = 1.0 | IF SMA_sell = 0.0 AND MACD_sell = 1.0 AND MO_sell = 0.0 OR RSI_sell = 1.0 AND LW_sell = 1.0 |
| 3 | [1.395, -0.124] | [2.442, -0.068] | IF SMA_buy = 0.0 OR sto_buy = 0.0 AND CCI_buy = 1.0 AND LW_buy = 1.0 | IF MACD_sell = 1.0 AND MO_sell = 1.0 AND PO_sell = 0.0 OR sto_sell = 0.0 AND RSI_sell = 0.0 AND LW_sell = 1.0 AND BB_sell = 1.0 |
| 4 | [1.432, -0.125] | [2.254, -0.054] | IF MACD_buy = 1.0 AND MO_buy = 0.0 OR RSI_buy = 0.0 | IF SMA_sell = 0.0 AND MACD_sell = 1.0 AND MO_sell = 1.0 OR sto_sell = 1.0 AND LW_sell = 1.0 AND BB_sell = 1.0 |

In Table 10, four strategies were generated as a part of the Pareto front. The results in the out-sample period here are superior to the in-sample period in a majority of the strategy with respect to both the objectives considered.

**Table 11. Strategies in In-Sample Period: 2013-14 Out-Sample Period: 2015**

| S.No | Sharpe Ratio & Drawdown (In-Sample) | Sharpe Ratio & Drawdown (Out-Sample) | Buy Strategy | Sell Strategy |
|---|---|---|---|---|
| 1 | [1.366, -0.118] | [-0.447, -0.126] | IF RSI_buy = 0.0 AND LW_buy = 0.0 | IF SMA_sell = 1.0 AND MACD_sell = 1.0 OR CCI_sell = 1.0 AND BB_sell = 1.0 |
| 2 | [1.155, -0.107] | [0.248, -0.076] | IF SMA_buy = 1.0 AND MO_buy = 1.0 AND PO_buy = 1.0 | IF MACD_sell = 1.0 OR RSI_sell = 0.0 AND CCI_sell = 1.0 AND LW_sell = 0.0 |

In Table 11, a pareto front of two strategies was generated. An interesting observation here is that both the buy strategies are completely concentrated on a single class of technical indicators.

## 6. Conclusions and future scope

The application of evolutionary computation algorithms in the finance industry – especially in the trading sector- is an incredibly fascinating area of research today. While risk-return trade-off is a standard area of study in multi-objective optimization, we introduce other unique combinations of objectives and suggest that the possibilities are endless and should be explored.

Furthermore, we introduce the idea of enforcing sensible constraints in rule generation to enhance their theoretical interpretability for the industry. We also emphasise on the importance of using rolling-based approaches and thus consider the effects of constantly evolving market conditions. Transaction costs should be strictly considered to make strategy generation as feasible as possible.

We observe that stable market conditions allow our strategies to perform incredibly well in the market. In future work, we recommend considering intra-day data (as technical indicators are best used here) and we strongly believe that using a rolling-based approach could induce well-performing strategies here. Finally, we encourage experimenting with other unique constraints and combinations of objectives as interesting areas of research. Further, reinforcement learning can be employed in conjunction with the multi-objective optimization algorithms to obtain better strategies.

**Appendix**

**Definitions of the Technical Indicators**

1. Simple Moving Average
2. MACD
3. Momentum Oscillator
4. Price Oscillator
5. Stochastic Oscillator
6. Relative Strength Index
7. CCI
8. Larry Williams

9. Bollinger Bands

These indicators are in the same order as mentioned in the chromosome structure. The first four indicators are Momentum Indicators, and the remaining are Reversal Indicators.

**Simple Moving Average**

The $SMA_n$ is given by calculating the average of the closing prices of the previous n days. We use two $SMA_n$ lines- $SMA_9$ and $SMA_{40}$ respectively to generate long and short signals.

$$SMA_n = \frac{\sum_{i=1}^{n} A_i}{N}$$

where $A_i$ refers to the closing price of the current day.

SMA_BUY=1 if $SMA_9(n-1) < SMA_{40}(n-1)$ and $SMA_9(n) > SMA_{40}(n)$

SMA_SELL=1 if $SMA_9(n-1) > SMA_{40}(n-1)$ and $SMA_9(n) < SMA_{40}(n)$

These signals try to indicate the formation of an upward and downward trend by comparing the average of recent prices with those collected over a longer period. It is therefore a Momentum indicator.

**The MACD Indicator**

The Moving Average Convergence Divergence Indicator is a momentum indicator that uses the difference between two $EMA_n$ (exponential moving averages) lines to capture trends. The MACD line is obtained by subtracting EMA26 from EMA12. This is followed by comparing crossovers between the MACD line and a Signal Line to generate Long/Short signals. The Signal Line is a smoothed version of the MACD line.

$$EMA_N(n) = \frac{2C(n)}{N+1} + \left(1 - \frac{2}{N+1}\right) * EMA_N(n-1)$$

where $C(n)$ represents the closing prices of the $n^{th}$ day.

MACD(n) = $EMA_{12}(n)$ - $EMA_{26}(n)$

$EMA_{12}(1)$ = $EMA_{26}(1)$ = C(1)

The Signal Line is given by 9-period exponentially smoothing the MACD line.

Signal Line(n) = 2MACD(n)/(9+1) + (1- 2/(9+1))Signal Line(n-1)

Signal Line(1) = MACD (1)

A long signal is generated on the day when the MACD line crosses above the Signal Line and a short signal is generated when the Signal Line crosses above the MACD line,

MACD_BUY=1 if MACD(n-1) < Signal Line(n-1) and MACD(n) > Signal Line(n)

MACD_SELL=1 if MACD(n-1) > Signal Line(n-1) and MACD(n) < Signal Line(n)

**Momentum Oscillator**

The Momentum Indicator is a simple indicator that just considers the difference in prices of the closing prices of the $n^{th}$ day and the $(n-10)^{th}$ day.

$$Momentum(n) = C(n) - C(n-10)$$

MomentumBuy(n) = 1 if Momentum(n-1) < 0 AND Momentum(n) > 0

MomentumSell(n) = 1 if Momentum(n-1) > 0 AND Momentum(n) < 0

It's a momentum indicator that captures upwards and downward trends to make trading decisions.

**Price Oscillator**

The Price Oscillator, similar to the MACD indicator, is based on a difference of EMA lines. However, we do not use a further smoothed version like the Signal Line used earlier to make trading decisions.

$$PO(n) = (EMA_{10}(n) - EMA_{20}(n))/EMA_{20}(n)$$

Like the Momentum Oscillator we generate long signals on the day PO(n) crosses zero upwards and short signals on the day PO(n) downwards. The Price Oscillator is also a momentum indicator.

**Stochastic Oscillator**

The Stochastic Oscillator is categorised as a reversal indicator. It compares closing prices to the range of prices the asset had for the time considered.

First let us consider a line K, where K is given by

$$K(n) = 100 * (C(n) - L14(n))/(H14(n) - L14(n))$$

where H14 and L14 are the highest high and lowest low prices (given by the High and Low columns in our OHLC data) in the last 14 days respectively.

We then have two more lines called D and $D_{slow}$ respectively given by

D(n) = SMA$_3$(K(n))

$D_{slow}$(n) = SMA$_3$(D(n))

where SMA$_3$ is the simple moving average of the previous three periods as mentioned earlier.

The Long and Short positions are taken as given by the rules given below-

SOBuy = 1 if D(n) < 20 AND $D_{slow}$(n) < 20 AND D(n-1) < $D_{slow}$(n-1) AND D(n) > $D_{slow}$(n)

SOSell = 1 if D(n) > 80 AND $D_{slow}$(n) > 80 AND D(n-1) > $D_{slow}$(n-1) AND D(n) < $D_{slow}$(n)

The thresholds 20% and 80% are taken from literature.

**Relative Strength Index**

The RSI indicator is used to identify overbought and oversold conditions in the market. It is a reversal indicator.

We start off by calculating the relative strength first.

$$RS(n) = ((AvgGain(n)/(AvgLoss(n)))$$

where $AvgGain = SMA14(max((C(n) - C(n - 1), 0))$ and

$$AvgLoss = SMA14(max((C(n - 1) - C(n), 0))$$

RSI(n) = 100 – (100/(1+RS(n)))

The long and short rules are:

RSIBuy=1 if RSI(n-1) < 30 and RSI(n) > 30

$$\text{RSISell} = 1 \text{ if RSI}(n-1) > 70 \text{ and RSI}(n) < 70$$

**Commodity Channel Index**

The CCI is again used to identify overbought and oversold conditions in the market. It can be calculated in the following steps.

First, we calculate the typical price of an asset given by:

$$TypicalPrice(n) = (C(n) + High(n) + Low(n))/3$$

We smoothen the typical prices by using 20-day averages given by $SMA_{20}(TypicalPrice(n))$.

Mean deviation(n) = $SMA_{20}(abs(SMA_{20}(TypicalPrice(n))-TypicalPrice(n)))$

where abs() represents the absolute value.

Finally, CCI is given by-

$$CCI(n) = \frac{(TypicalPrice(n) - SMA20(TypicalPrice(n)))}{(0.015 * Mean\ Deviation(n))}$$

The long and short rules for CCI are:

$$\text{CCIBuy} = 1 \text{ if CCI}(n-1) < 100 \text{ and CCI}(n) > 100$$

$$\text{CCISell} = 1 \text{ if CCI}(n-1) > -100 \text{ and CCI}(n) < -100$$

**Larry Williams**

The Larry Williams Indicator is very similar to the Stochastic Oscillator. It is a reversal indicator. It is given by-

$$LW(n) = 100 * (C(n) - H14(n))/(H14(n) - L14(n))$$

where L14 and H14 are the same as that mentioned in the Stochastic Oscillator section.

The LW rules for long and short signals are-

$$\text{LWBuy} = 1 \text{ if LW}(n-1) < -80 \text{ and LW}(n) > -80$$

$$\text{LWSell} = 1 \text{ if LW}(n-1) > -20 \text{ and LW}(n) < -20$$

**Bollinger Bands**

Bollinger Bands are price envelopes developed by John Bollinger. They involve an upper band and lower band and are used to determine whether prices are high or low. This is a reversal indicator.

These bands are three standard deviations above and below the simple moving average of a period of 20 days.

$$UpperBand = SMA_{20}(n) + 3*StdDev(n)$$

$$LowerBand = SMA_{20}(n) - 3*StdDev(n)$$

The rules are

$$BBandBuy = 1 \text{ if } C(n-1) < LowerBand(n-1) \text{ and } C(n) > LowerBand(n)$$

$$BBandSell = 1 \text{ if } C(n-1) > UpperBand(n-1) \text{ and } C(n) < UpperBand(n)$$